\def\year{2020}\relax
\documentclass[letterpaper]{article} 
\usepackage{aaai20}  
\usepackage{times}  
\usepackage{helvet} 
\usepackage{courier}  
\usepackage[hyphens]{url}  
\usepackage{graphicx} 

\usepackage{amsmath}
\usepackage{amssymb}
\usepackage{subfigure}
\usepackage{booktabs}
\usepackage{multirow}
\usepackage{longtable}

\def\UrlFont{\rm}  
\usepackage{graphicx}  
\title{Generative Adversarial Networks for Video-to-Video Domain Adaptation}
\author{Jiawei Chen\textsuperscript{\rm}, Yuexiang Li\textsuperscript{\rm}\thanks{Corresponding author}, Kai Ma\textsuperscript{\rm}, Yefeng Zheng\textsuperscript{\rm}\\ 
\textsuperscript{\rm}YouTu Lab, Tencent, Shenzhen, China\\ 
\{jiaweichen, vicyxli, kylekma, yefengzheng\}@tencent.com 
}

\begin{document}

\maketitle

\begin{abstract}
    Endoscopic videos from multicentres often have different imaging conditions, e.g., color and illumination, which make the models trained on one domain usually fail to generalize well to another. Domain adaptation is one of the potential solutions to address the problem. However, few of existing works focused on the translation of video-based data. In this work, we propose a novel generative adversarial network (GAN), namely VideoGAN, to transfer the video-based data across different domains. As the frames of a video may have similar content and imaging conditions, the proposed VideoGAN has an X-shape generator to preserve the intra-video consistency during translation. Furthermore, a loss function, namely color histogram loss, is proposed to tune the color distribution of each translated frame. Two colonoscopic datasets from different centres, i.e., CVC-Clinic and ETIS-Larib, are adopted to evaluate the performance of domain adaptation of our VideoGAN. Experimental results demonstrate that the adapted colonoscopic video generated by our VideoGAN can significantly boost the segmentation accuracy, i.e., an improvement of 5\%, of colorectal polyps on multicentre datasets. As our VideoGAN is a general network architecture, we also evaluate its performance with the CamVid driving video dataset on the cloudy-to-sunny translation task. Comprehensive experiments show that the domain gap could be substantially narrowed down by our VideoGAN.
\end{abstract}

\section{Introduction}
\begin{figure}[!t]
    \begin{center}
        \includegraphics[width=0.95\linewidth]{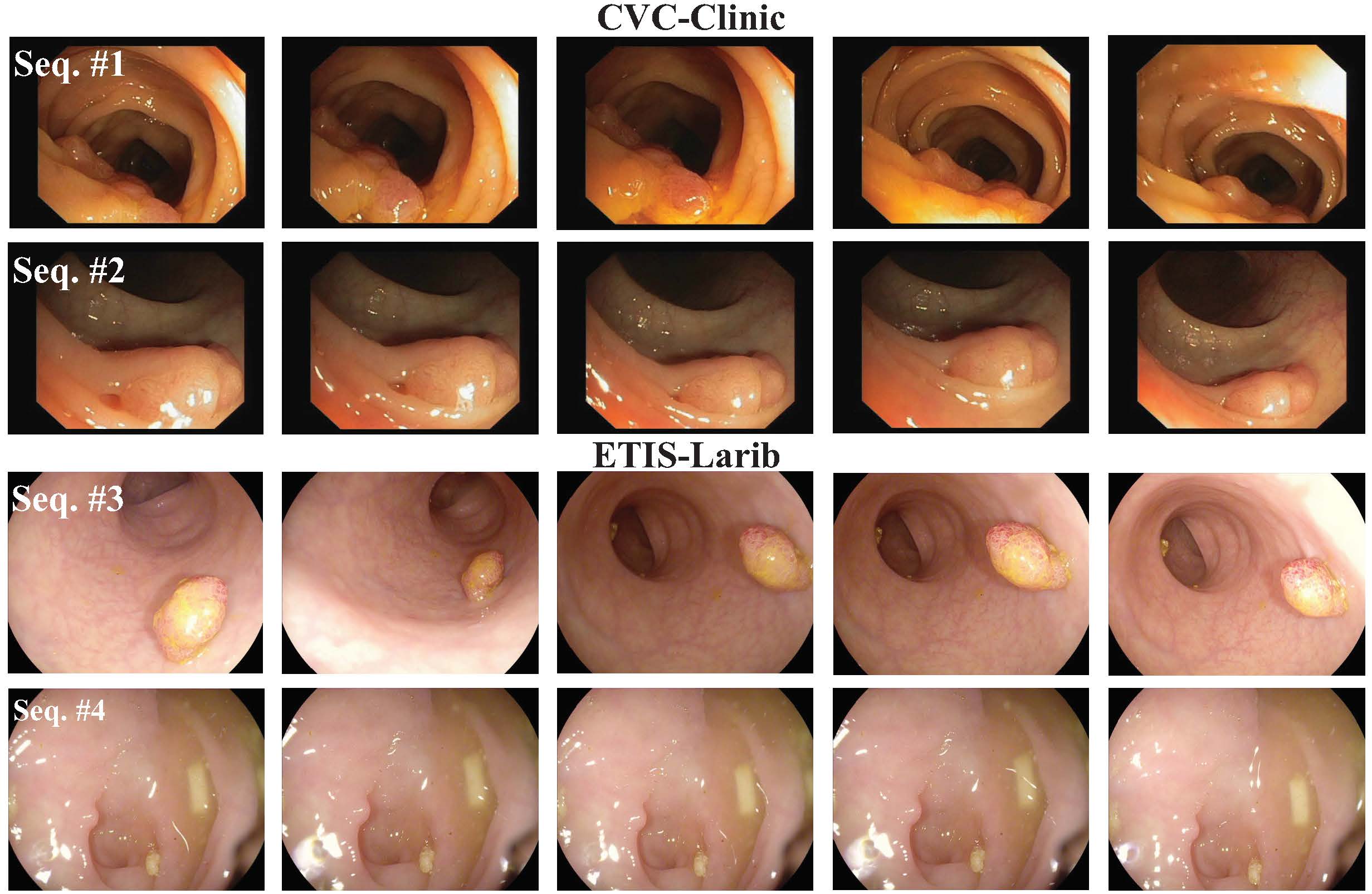}
    \end{center}
    \caption{Examples of colonoscopy videos from multicentres. The colonoscopic images from CVC-Clinic and ETIS-Larib databases have different distributions of color and illumination.}
    \label{fig1:long}
    \label{fig1:onecol}
\end{figure}

The colorectal and stomach cancers are the leading causes of worldwide cancer deaths in 2018, accounting for 9.2\% and 8.2\% of total cancer deaths, respectively \cite{BrayF01}. The endoscopy is the primary imaging modality for screening and diagnosis of these cancers and over 100 million endoscopy exams are annually performed. However, the traditional screening approach requires specialized physicians to visually analyze extensive endoscopic videos, which is extremely laborious and suffers from various problems, e.g. inter-observer variations. With the recent development of deep learning, an increasing number of studies tried to develop computer-aid diagnosis (CAD) system for endoscopic images. For example, Zhang et al. \cite{no_1} proposed a regression-based convolutional neural network (R-CNN) pipeline for the automated detection of polyps during colonoscopy. Yu et al. \cite{no_3} developed an offline and online 3D deep learning integration framework by leveraging the 3D fully convolutional network (3D-FCN) to tackle the problem of colorectal polyp detection. Although the existing studies significantly improved the diagnosis performance of CAD systems, the generality of the established models was merely investigated as data from single source domain was often used. The endoscopic images from multicentres usually have different imaging conditions such as color distribution and illumination effects, because imaging devices and imaging parameter settings are not well standardized. As shown in Fig.~\ref{fig1:long}, the colonoscopic video frames captured by the CVC-Clinic centre \cite{VD01} is warm-toned (higher red intensity) compared to the ones from the ETIS-Larib centre \cite{no_9}. Such variations would make models trained on one domain fail on another.

Domain adaptation is one of the potential solutions to address the problem of variations of imaging conditions among multi-domains. Although the area of domain adaptation has been extensively studied, most of the existing works focus on the image-to-image domain adaptation \cite{Isola01,no_4}, which may not be the optimal solution for video-based endoscopic data. As shown in Fig.~\ref{fig1:long}, the frames in the same endoscopic video have similar color distributions. To this end, specific constraints on video data, i.e., intra-video color consistency, also need to be applied during the domain adaptation process.

In this paper, we propose a novel generative adversarial network, namely VideoGAN, for domain adaptation of multicentre endoscopic videos. Our VideoGAN addresses two key problems that are overlooked by the current GAN-based translation approaches: 1) video content distortions, and 2) intra-video color inconsistency. To maintain the content and intra-video color consistencies during the domain adaptation process, the proposed VideoGAN model incorporates an X-shape generator that simultaneously translates two frames, i.e., the reference and source frames, to the same mode in the target domain. The mode defined for video-to-video domain adaptation is a video clip containing similar contents, which is totally different from that of the image-to-image translation, i.e., a single image. Meanwhile, we also want to preserve the relative shift in color distributions between the source and reference frames, e.g., if the source frame is brighter than the reference frame, this trend should be preserved after translation. To this end, we propose a novel color histogram loss that calculates the difference between two histograms of source and reference frames.

The proposed VideoGAN is evaluated on publicly available colonoscopic video datasets. The results illustrate that our VideoGAN overcomes the distortion problem of standard CycleGAN and yields elegant adapted videos, which significantly boosts the segmentation accuracy of colorectal polyps. Furthermore, as our VideoGAN is a general network architecture, we also evaluate its performance on natural video datasets. The experiments demonstrate that the proposed VideoGAN is suitable for natural video adaptation as well.

\section{Related Work}
Few existing works focus on domain adaptation of videos. Two closely related areas to this work are image-to-image translation and unsupervised domain adaptation. Hence, we briefly summarize those two categories of works in this section.

\subsection{Image-to-image translation}
Since firstly proposed by Goodfellow et al. \cite{Goodf01}, GAN attracts increasing attentions from the community. Many tasks have been successfully tackled by GAN and its variations \cite{ChangH01,ChenY01,MaS01}, such as image synthesis \cite{WangT01}, super-resolution \cite{Ledig01} and image translation \cite{Isola01,no_4,ChenY01,MaS01}. Image-to-image translation aims at constructing a pixel-to-pixel mapping between two domains. A representative method is the conditional GAN \cite{Isola01}, which shows a strategy to learn such translation mapping with a conditional setting to capture structure information. However, it requires paired cross-domain images as training data, which are often difficult to aquire. To perform unpaired image-to-image translation, several works \cite{no_4,Kim01,Yi01} have been recently proposed. Those works introduce a cycle consistency loss to loose the confinement of paired training images. The proposed VideoGAN adopts the idea of cycle consistency and extends the framework for unpaired video translation. We notice that there are some studies on the topic of video synthesis \cite{Chen01,WangTC01}. However, those works still require paired training data to perform video-to-video translation.

\begin{figure*}[!htb]
    \begin{center}
        \includegraphics[width=0.75\linewidth]{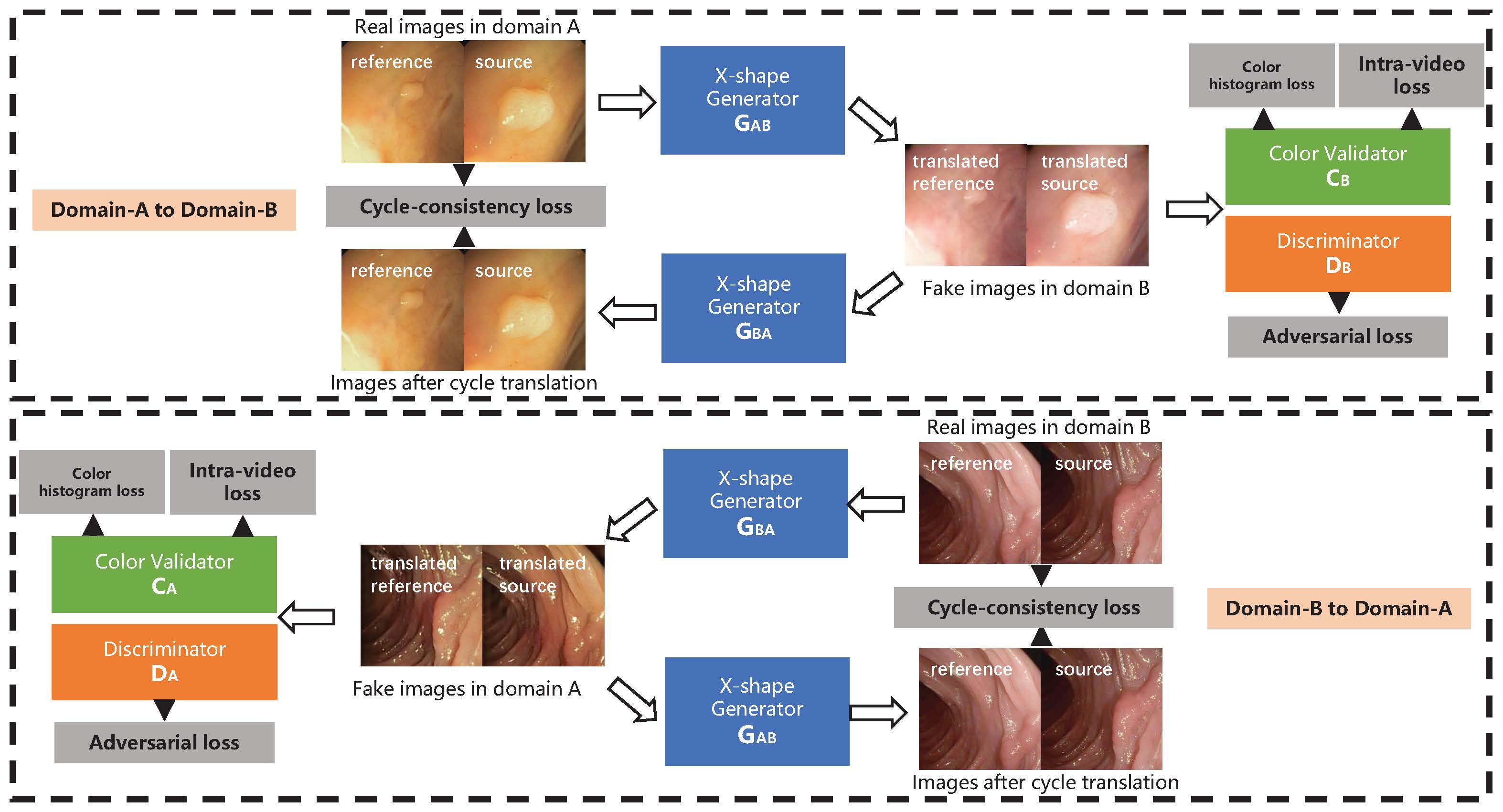}
    \end{center}
    \caption{The overview of our VideoGAN. The proposed VideoGAN translates endoscopic videos between the domain-A (CVC-Clinic) and domain-B (ETIS-Larib). The generators ($G_{AB}$ and $G_{BA}$) are supervised by a cycle-consistency loss, an adversarial loss, an intra-video loss and a color histogram loss. To preserve the intra-video consistency, a reference frame is sent with the source frame together as input to the X-shape generator.}
    \label{fig:short}
\end{figure*}

\subsection{Unsupervised domain adaptation}
Apart from image-to-image translation, another area related to our work is the unsupervised domain adaptation (UDA). The UDA aims to close the gap, such as the color distribution, between the source and target domains without target sample annotations. One common choice for domain adaptation is to establish a mapping in the feature subspace between two different domains \cite{Sun01,GaninY01}. For example, Sun et al. \cite{Sun01} proposed the CORrelation Alignment (CORAL) method for unsupervised domain adaptation, which minimizes domain shift by aligning the second-order statistics of source and target distributions. In more recent studies, researchers tried to adopt GANs to seek an optimal feature space to build the mapping between two domains \cite{Hoffman01,SankaranarayananS01,HuangS01,Chen_2019_ICCV}. Hoffman et al. \cite{Hoffman01} developed an approach, called cycle-consistent adversarial domain adaptation (CyCADA), to guide transfer between domains according to a discriminatively trained network. The approach alleviated the divergence problem by enforcing consistency of the relevant semantics before and after adaptation. Recently, the GAN-based domain adaptation is also widely used in the area of person re-identification (re-ID) \cite{no_5,Wei01} and medical image processing, e.g., the color normalization of histopathological slices \cite{no_6}, intensity standardization of magnetic resonance images \cite{no_7} and cross-modality adaptation \cite{Zhang01}. However, most of the existing GAN-based domain adaptation approaches require strong prior-knowledge, e.g., pixel-wise annotations \cite{Zhang01,HuangS01}, which limits their applications for general tasks. {\itshape Our VideoGAN looses the requirements of prior-knowledge and achieves domain adaptation without any supervision signal.}

\section{Proposed Method}
In this section, we introduce VideoGAN in details. We first revisit the principle of CycleGAN and then introduce the proposed X-shape generator, color validator and color histogram loss.

\subsection{Revisit of CycleGAN}
CycleGAN has two paired generator-discriminator modules, which are capable of learning two mappings, i.e., from domain A to domain B \{$G_{AB}$, $D_B$\} and vice versa \{$G_{BA}$, $D_A$\} . The generators ($G_{AB}$, $G_{BA}$) translate images between the source and target domains, while the discriminators ($D_A$, $D_B$) aim to distinguish the real and translated data. Thereby, the generators and discriminators are gradually updated during this adversarial competition.

The original CycleGAN is supervised by two losses, i.e., adversarial loss ($\mathcal{L}_{adv}$) and cycle-consistency loss ($\mathcal{L}_{cyc}$). The adversarial loss encourages local realism of the translated data. Taking the translation from domain B to domain A as an example, the adversarial loss can be written as:

\begin{equation}
    \begin{aligned}
        \mathcal{L}_{adv}(G_{BA},D_{A})=\mathbb{E}_{x_{A}\sim p_{x_{A}}}\left [ (D_{A}(x_{A})-1)^{2} \right ] \\
        +\mathbb{E}_{x_{B}\sim p_{x_{B}}}\left [ (D_{A}(G_{BA}(x_{B})))^{2} \right ]
    \end{aligned}
\end{equation}
where $p_{x_{A}}$ and $p_{x_{B}}$ denote the sample distributions of domain A and B, respectively; $x_{A}$ and $x_{B}$ are samples from domain A and B, respectively.

The cycle-consistency loss ($\mathcal{L}_{cyc}$) tackles the problem of deficient paired training data. The idea behind the cycle-consistency loss is that the translated data from the target domain can be exactly converted back to the source domain, which can be expressed as:

\begin{equation}
    \begin{aligned}
        \mathcal{L}_{cyc}(G_{BA},G_{AB})=\mathbb{E}_{x_{A}\sim p_{x_{A}}}\left [ \left \| G_{BA}(G_{AB}(x_{A}))-x_{A} \right \|_{1} \right ] \\
        +\mathbb{E}_{x_{B}\sim p_{x_{B}}}\left [ \left \| G_{AB}(G_{BA}(x_{B}))-x_{B} \right \|_{1} \right ]
    \end{aligned}
\end{equation}
where the $\mathcal{L}_{1}$ loss is adopted in $\mathcal{L}_{cyc}$ in our VideoGAN.

{\bf Problems of using CycleGAN in video-to-video domain adaptation:} First, due to the intrinsic ambiguity with respect to geometric transformations of CycleGAN \cite{Zhang01}, content distortion may exist in the translated results produced by CycleGAN.  Second, as the CycleGAN separately transfers each frame of the videos, the frames may be mapped to different modes of the target domain. Such problems violate the intuition of domain adaptation of videos, but they have not been addressed in existing methods.



\subsection{VideoGAN}
We propose the VideoGAN to solve the aforementioned problems of using CycleGAN in the video-to-video domain adaptation. Fig.~\ref{fig:short} illustrates the workflow of our VideoGAN. The proposed VideoGAN shares a similar cyclic processing procedure as CycleGAN. To preserve the intra-video consistency and prevent the content distortion, however, we set a reference frame with the source frame as a dual-input to the generator, which also generates two corresponding frames in the target domain. We name the generator as X-shape generator because the dual-in-dual-out and the encode-decode properties formulate the network structure in an X-shape. Beyond the regular discriminator that distinguishes real/fake images, we develop another color validation module (referred as color validator for convenience) that tunes the color distribution of the source frame according to the paired reference frame. Supervised by the proposed losses (adversarial loss, cycle-consistency loss, color histogram loss and intra-video loss), VideoGAN is trained in an end-to-end fashion that follows the protocol proposed in \cite{no_4}.

\subsubsection{X-shape generator.}
In order to maintain the consistency, i.e., content and color, of generated frames belonging to the same video, the proposed X-shape generator simultaneously translates two images, i.e., source and reference frames, from the source domain to the target domain. As there are multiple modes in the target domain, to bring the generated frames to a certain mode, the reference frame acts as an anchor to pull other source frames to be into the same mode. Hence, the reference frame (anchor) should be invariant during the translation of the whole video. In our experiments, the first frame of each video\footnote{For a long video, we truncate it into a set of clips based on scene changes. And, each video clip has a consistent scene setting.} is chosen as the reference frame, which is paired with each remaining frame of the corresponding video and sent to the generator as the input. Consequently, the reference frame appears multiple times during training and gains more weights than other source frames, which can help determine the marching direction of the rest video frames. Furthermore, the extracted features of the reference and source frames are embeded into a new feature space by a dense fusion block, where the fusion process can be seen as adding mutual watermarks to both frames. The content distortions caused by the bijective geometric transformation and its inverse operation \cite{Zhang01} can be prevented, since they may scramble the latent watermarks and do not maintain the cycle consistency any more.

\begin{figure}[htb]
    \begin{center}
        \includegraphics[width=0.95\linewidth]{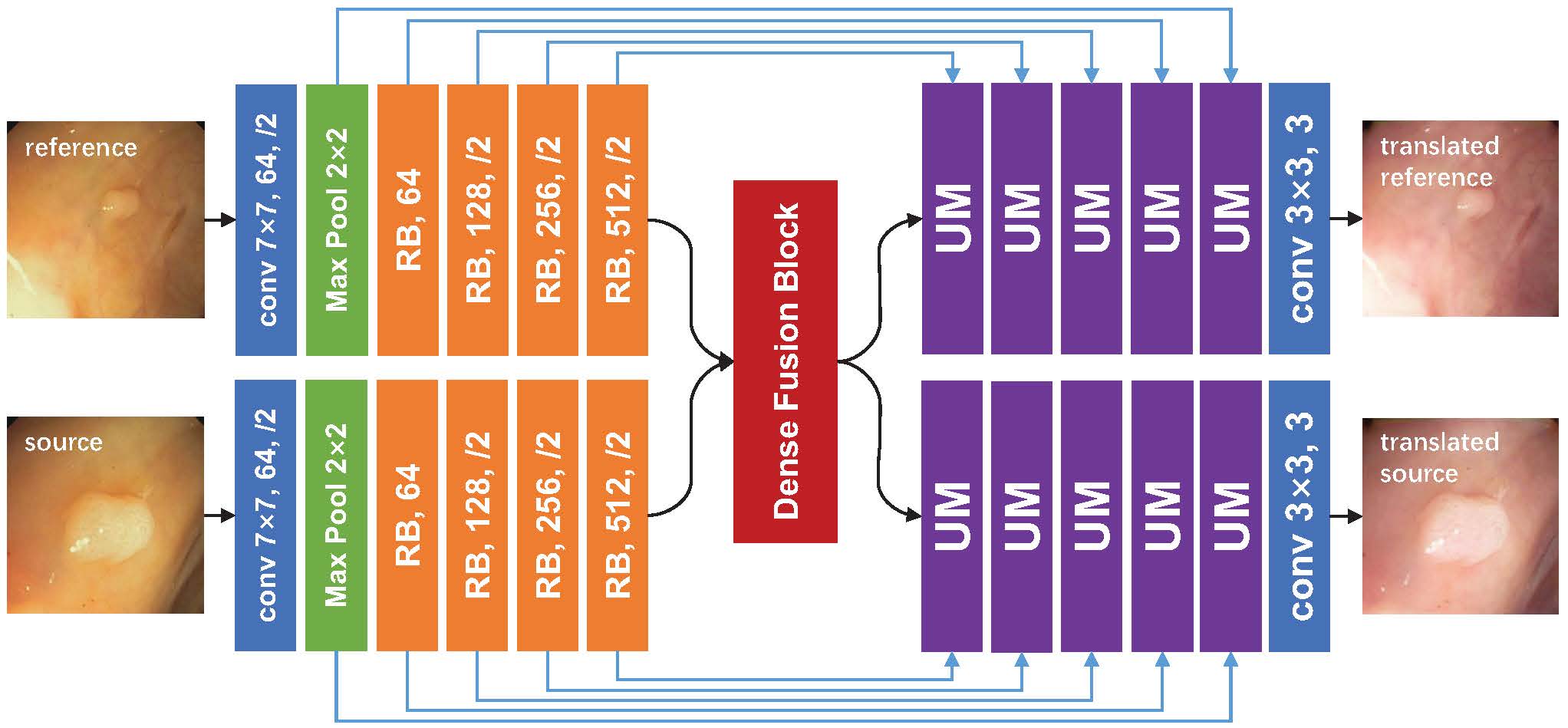}
    \end{center}
    \caption{The proposed X-shape generator. The source and reference frames have separate encoder-decoder modules. The size of source and reference frames is $256 \times 256 \times 3$. The extracted features from the two frames are fused in the feature space by a dense fusion block. The blue, green, orange, red and purple rectangles represent the convolution layer, max-pooling layer, residual block, dense fusion block and upsampling layer, respectively.}
    \label{fig3:long}
    \label{fig3:onecol}
\end{figure}

Fig.~\ref{fig3:long} presents the network architecture of our X-shape generator. The blue, green, orange, red and purple rectangles represent the convolution layer, max-pooling layer, residual block (RB) \cite{He01}, dense fusion block and upsampling module (UM, consisting of a 3 $\times$ 3 convolution layer and an upsampling layer), respectively. As shown in Fig.~\ref{fig3:long}, the source and reference frames are processed by independent encoders and decoders. The extracted high-level features are fused by the dense fusion block to preserve the intra-video consistency and alleviate the problem of content distortion. The short-cut connection \cite{Ronneberger01} is employed to facilitate the feature information flowing between the encoders and decoders.

\begin{figure}[htb]
    \begin{center}
        \includegraphics[width=0.9\linewidth]{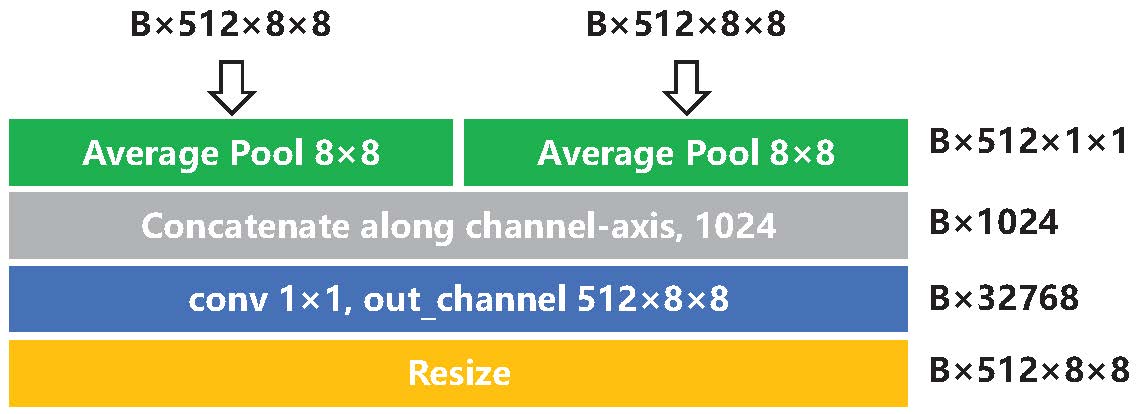}
    \end{center}
    \caption{The proposed dense fusion block. It uses the average pooling layer (green rectangles) and 1$\times$1 convolution to fuse the high-level features of the source and reference frames.}
    \label{fig4:long}
    \label{fig4:onecol}
\end{figure}

{ \bf Dense fusion block.}
The dense fusion block ensures that the source frames are mapped to the same mode in the target domain, and prevents content distortions by embedding content information of the reference frame to the source frames. Fig.~\ref{fig4:long} shows the architecture of our dense fusion block. The green, gray and yellow rectangles represent the average pooling, concatenation and resizing layers, respectively. Naive fusion approaches, e.g., stacking two images as different input channels or fusing the feature maps generated at the middle stages of encoders, suffer from the spatial information interference between the paired frames. Hence, the 8$\times$8 feature maps from the source and reference frames are average-pooled to 1$\times$1 size before the fusion. The symbol B in Fig.~\ref{fig4:long} stands for the batchsize and 512 is the number of feature maps. Inspired by the dense upsampling convolution \cite{Wang01}, a 1$\times$1 convolution with $512\times8\times8$ channels is used to fuse the information of source and reference frames. The fused $32768\times1\times1$ feature maps are then resized back to size of $512\times8\times8$ and sent to the decoders.

\subsubsection{Color validator.}
Color validators ($C_{A}$, $C_{B}$) in our VideoGAN perform two tasks. First, they regulate the color variation trend of the translated frames with the color histogram loss. For example, suppose the source frame is brighter than the reference frame, we want to maintain the relationship after mapping the source frame to the target domain. Second, they act as secondary discriminators assessing the real/fake identities of the paried frames with an intra-video loss. In this setting, the actual samples input to the color validators are any two frames of the same video from target domain. Therefore, to trick such color validators, the generators latently maintain the long-term intra-video consistency. The color validator of our VideoGAN takes the two translated frames as the input and yields a $1\times15$ ($C^{hist}$) and a $1\times1$ ($C^{i.v.}$) vector as predictions to calculate the color histogram loss and intra-video loss, respectively.

{ \bf Color histogram loss.}
The reference frame in our VideoGAN provides the identity information as well as the relative color distribution, which can be used to tune the color of source frames. Assuming the histogram of each of RGB channels can be expressed as $hist_{c}$ ($c \in  R, G, B$), the relative color distribution ($hist_{rcd}$) can be defined as:

\begin{equation}
    \begin{aligned}
        hist_{rcd} = cat(hist_{R}^{ref}, hist_{G}^{ref}, hist_{B}^{ref}) \\
        - cat(hist_{R}^{src}, hist_{G}^{src}, hist_{B}^{src})
    \end{aligned}
\end{equation}
where the $hist^{src}$ and $hist^{ref}$ refer to the histograms of source and reference frames, respectively; $cat(.)$ is the concatenation operation.

The translated frames are expected to maintain the information of relative color distribution after domain adaptation. Hence, we calculate a 5-bin histogram for each of RGB channels and form a 15-bin $hist_{rcd}$ as ground truth to supervise the VideoGAN. The color histogram loss ($\mathcal{L}_{hist}$) can thereby be written as:

\begin{equation}
    \begin{split}
        \mathcal{L}_{hist}(G_{BA}, C_{A}^{hist})=&\| C_{A}^{hist}(G_{BA}(x_{B}^{src}), G_{BA}(x_{B}^{ref}))\\
        &-hist_{rcd}(x_{B}^{src}, x_{B}^{ref})  \|_{1}
    \end{split}
\end{equation}
where the $\mathcal{L}_{1}$ loss is adopted in $\mathcal{L}_{hist}$ in our VideoGAN; $x_{B}^{src}$ and $x_{B}^{ref}$ are the source and reference frames from domain B, respectively.

{ \bf Intra-video loss.}
The intra-video loss ($\mathcal{L}_{i.v.}$) enables our VideoGAN to maintain the intra-video consistency for lengthy videos, such as the ones in CamVid dataset.\footnote{http://mi.eng.cam.ac.uk/research/projects/VideoRec/CamVid} The color validator distinguishes the translated paired frames from the actual frames selected from the same video of the target domain, which enforces the generator to consider long-term intra-video consistency. The intra-video loss has the same form as the adversarial loss ($\mathcal{L}_{adv}$), which can be denoted as $\mathcal{L}_{i.v.}(G_{BA},C_{A}^{i.v.})$.

\subsubsection{Discriminator.}
In consistent with the standard CycleGAN, the proposed VideoGAN has discriminators ($D_{A}$, $D_{B}$) to validate the quality of each translated frame. The SegNet \cite{BadrinarayananV01} is adopted to provide pixel-wise prediction of real or fake, rather than to classify the whole frame or sub-image. Such an approach encourages the discriminators to take more information into account while making real/fake decisions.

\subsubsection{Objective.}
Apart from the cycle-consistency, adversarial, color histogram and intra-video losses defined above, we also involve the identity constraint \cite{no_5}, i.e., $\mathcal{L}_{idt}$, as an auxiliary loss. The target domain identity constraint regularizes the generator to be the identity matrix on samples from target domain. Finally, our full objective function is summarized as:

\begin{equation}
    \begin{split}
        \mathcal{L}(G_{BA}, G_{AB}, D_{A}, D_{B}, C_{A}, C_{B})=\mathcal{L}_{adv}(G_{BA},D_{A}) \\
        + \mathcal{L}_{adv}(G_{AB}, D_{B}) + \mathcal{L}_{hist}(G_{BA}, C_{A}^{hist}) \\
        + \mathcal{L}_{hist}(G_{AB}, C_{B}^{hist}) + \mathcal{L}_{i.v.}(G_{BA},C_{A}^{i.v.}) \\
        + \mathcal{L}_{i.v.}(G_{AB},C_{B}^{i.v.}) + \mathcal{L}_{cyc}(G_{BA}, G_{AB}) \\
        + \mathcal{L}_{idt}(G_{BA}, G_{AB})\\
    \end{split}
\end{equation}

The optimization of $\mathcal{L}_{cyc}$ and $\mathcal{L}_{adv}$ is in an alternative manner following the protocol proposed in \cite{no_4}. The $\mathcal{L}_{idt}$ is updated at the same time of $\mathcal{L}_{cyc}$, while the $\mathcal{L}_{hist}$ and $\mathcal{L}_{i.v.}$ are updated with the $\mathcal{L}_{adv}$ optimization.

\section{Experiments}
This section evaluates and discusses the performance of our VideoGAN. Suffering from the problem of multicentres, the conditions of endoscopic videos often have wide variations, which dramatically degrade the generalization of a computer-aid diagnosis system trained on a single domain. In this section, we begin with tackling the multicentre problem for endoscopic videos using our VideoGAN. As the proposed VideoGAN is a general network architecture, we then evaluate its domain adaptation performance on the CamVid video dataset.

\subsubsection{Datasets.}
The publicly available colonoscopic video datasets, i.e., CVC-Clinic\footnote{https://polyp.grand-challenge.org/CVCClinicDB/} \cite{VD01} and ETIS-Larib\footnote{https://polyp.grand-challenge.org/EtisLarib/} \cite{no_9}, are selected for our experiments. The CVC-Clinic dataset is composed of 29 sequences with a total of 612 images. The ETIS-Larib consists of 196 images, which can be manually separated to 29 seuqences as well. Those short videos are extracted from the colonoscopy videos captured by different centres using different endoscopic devices. All the frames of the short videos contain polyps. The dataset providers annotate the pixel-wise ground truth covering the polyps for data users. Sample frames of the two datasets are shown in Fig.~\ref{fig1:long}.

\subsubsection{Training details.}
The proposed VideoGAN is implemented using PyTorch. The Adam solver \cite{Kingma01} with betas = (0.5, 0.999) is adopted for the optimization of VideoGAN. The network is trained with a mini-batch size of 1 on one GPU (Tesla P40 with 24 GB memory). The initial learning rate is set to 0.0002. The proposed VideoGAN yields visually satisfactory translated frames after 200 training epochs.

\subsubsection{Experiment settings.}
Given two datasets (A, B), our goal is to narrow down the gap between them not only in terms of visual perception i.e., elegant translated results, but also the representation in feature space, i.e., improvement of the robustness of models. We present the video translation results to evaluate the former factor. For the latter one, the learning via translation framework \cite{LiuMY01,no_5} is adopted for the evaluation. The framework consists of two steps, i.e., cross-domain image translation for training data creation, and supervised feature learning for a specific task, e.g., polyp segmentation. Accordingly, we evaluate VideoGAN in two scenarios: {\bf transfer learning} (CVC-Clinic $\rightarrow$ ETIS-Larib) and {\bf data augmentation} (ETIS-Larib $\rightarrow$ CVC-Clinic), respectively. In transfer learning, the target domain data has no labels. We use the target domain data to train CycleGAN and VideoGAN to translate the source domain data to the target domain. The translated source domain data are then used to train a segmentation network for the transfer learning on target domain. In data augmentation, the translated source domain data are added to the labeled target domain data to train a segmentation network.



{ \bf Evaluation criterion.} The Dice coefficient, which measures the spatial overlap index between the segmentation result and ground truth, is adopted as the metric to assess the accuracy of colorectal ployp segmentation.

{ \bf Baselines overview.} Based on the experiment settings, several unpaired image-to-image translation frameworks, i.e., G(.), are taken as baselines for the performance evaluation, including CycleGAN \cite{no_4}, UNIT \cite{LiuMY01} and DRIT \cite{Lee01}. We also compare VideoGAN with a direct transfer approach, which directly takes the source domain data for training without any adaptation. Note that the recent proposed GANs for image-based domain adaptation, e.g., SPGAN \cite{no_5}, PTGAN \cite{Wei01} and AugGAN \cite{HuangS01}, are not involved for comparison, due to the strong prior-knowledge used in those approaches. SPGAN used the prior-knowledge that the ID sets of different re-ID domains used in their experiments are totally different from each others. PTGAN required coarse segmentation results to distinguish foreground and background areas. AugGAN added a segmentation subtask to the CycleGAN-based framework, which required pixel-wise annotations. The use of prior-knowledge degrades the generalization of those GANs, which are only suitable for the domains fulfilling the specific requirements.

\subsection{Visualization of domain adaptation results}
\begin{figure}[!tb]
    \begin{minipage}[t]{0.49\linewidth}
        \centering
        { \scriptsize \bf CVC-Clinic $\rightarrow$ ETIS-Larib}
        \includegraphics[height=1.5in]{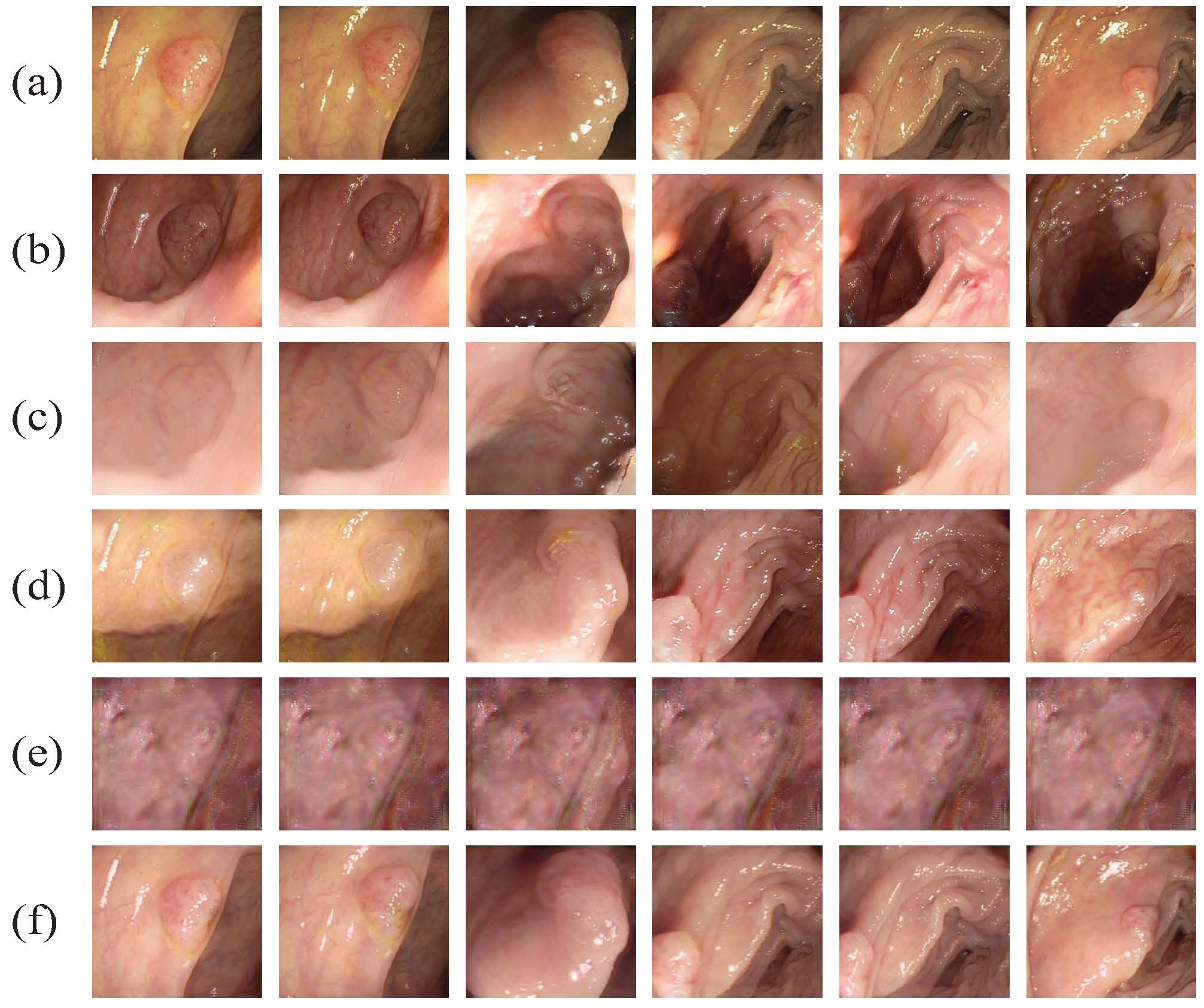}
    \end{minipage}
    \begin{minipage}[t]{0.49\linewidth}
        \centering
        { \scriptsize \bf ETIS-Larib $\rightarrow$ CVC-Clinic}
        \includegraphics[height=1.5in]{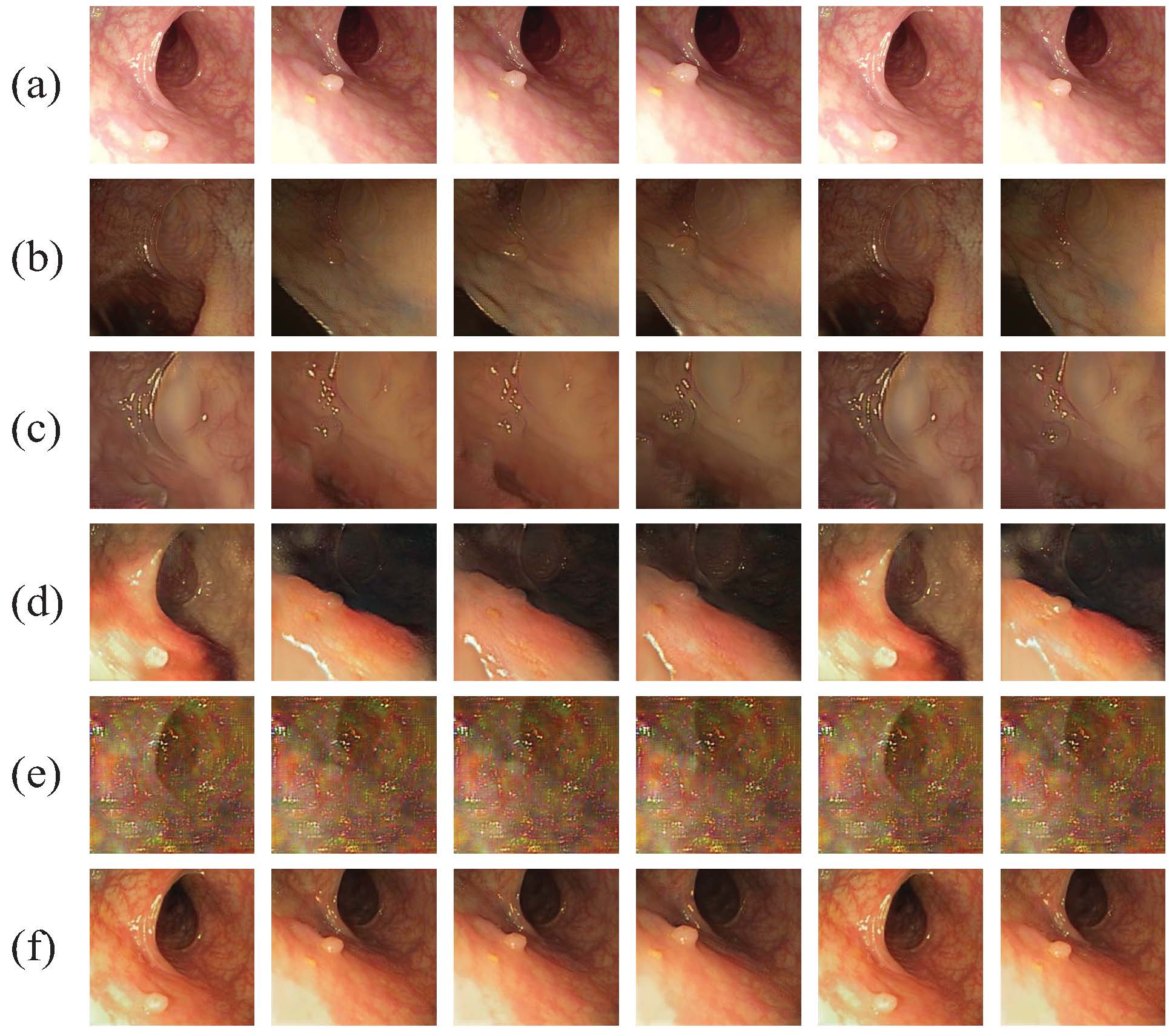}
    \end{minipage}
    \caption{The domain adaptation videos cross CVC-Clinic and ETIS-Larib domains. (a) The original videos in source domains. (b)-(e) are the domain adaptation videos produced by UNIT, DRIT, CycleGAN, and CycleGAN with stacked input, respectively. (f) The domain adaptation videos produced by VideoGAN.} \label{fig5:cross_domains}
\end{figure}

The domain adaptation results between CVC-Clinic and ETIS-Larib domains are shown in Fig.~\ref{fig5:cross_domains}, which illustrates two main problems of existing image-to-image translation approaches (UNIT \cite{LiuMY01}, DRIT \cite{Lee01} and CycleGAN \cite{no_4}), i.e., the content distortion and random color variation. The CycleGAN taking the source and reference frames as two input channels is also adopted for comparison. Due to the interference of spatial information of source and reference frames, the CycleGAN with stacked input fails to generate meaningful translation results. As Fig.~\ref{fig5:cross_domains} shows, our VideoGAN can simutaneously maintain the video contents and the trend of color variation throughout the video.

{ \bf Quantitative analysis of content distortions.} To quantitatively evaluate the content distortion degrees, we invite an experienced senior physician to annotate the polyps in the original ETIS-Larib videos and the ones translated by CycleGAN and our VideoGAN, respectively. To evaluate the gap between the physician and the ETIS-Larib experts, the Dice coefficient between new annotations on the original ETIS-Larib videos and the ground truths is measured.

As Table~\ref{tab:content_distortions} lists, due to inter-observer variability, the physician achieves a Dice coefficient of 82.73\% on the original ETIS-Larib dataset. The annotation accuracy of physician on translated videos should thereby be close to 82.73\%, if the contents are preserved through the translation. However, the Dice coefficient drops to 41.39\% while the physician annotating the video translated by CycleGAN. Due to content distortion, a polyp may totally disappear or move to a different location in some translated images of CycleGAN, resulting in zero Dice coefficient, which deteriorates the overall Dice coefficient significantly. On the contrary, since the proposed VideoGAN well maintains video contents before and after the video translation, the physician achieves a 80.37\% Dice coefficient on the videos translated by our VideoGAN.

\begin{table}[!htb]
    \caption{Dice coefficients between new annotations on different ETIS-Larib videos and the ground truth.}\label{tab:content_distortions}
    \begin{center}
        \begin{tabular}{l|c|c|c}
            \hline
            { }              & {\bfseries Original} & {\bfseries CycleGAN} & {\bfseries VideoGAN} \\
            \hline\hline
            { \bf Dice (\%)} & 82.73                & 41.39                & 80.37                \\
            \hline
        \end{tabular}
    \end{center}
\end{table}

\subsection{Evaluation of learning via translation}
As aforementoined, we conduct experiments to evaluate the proposed VideoGAN from two aspects, i.e., transfer learning and data augmentation.

\subsubsection{Transfer learning.} The size of ETIS-Larib dataset is extremely small (196 frames), which is insufficient to train a deep learning network. A common solution to this problem is the transfer learning, i.e., using additional related dataset, e.g. CVC-Clinic, as the training set for model learning. We evaluate the segmentation performance of FCNs, trained with CVC-Clinic videos translated by different GANs, on the ETIS-Larib dataset. ResUNet-50 \cite{He01,Ronneberger01} is chosen as the backbone for FCNs because it shows excellent performance on many segmentation tasks. Table~\ref{tab1} lists the polyp segmentation accuracies of FCNs trained with different training sets.

\begin{table}[!h]
    \caption{Comparison of Dice coefficients of polyp segmentation yielded by different domain adaptation methods on ETIS-Larib. The best result is in {\bf bold}.}\label{tab1}
    \begin{center}
        \begin{tabular}{l|c}
            \hline
            { }                              & { \bf Dice (\%)}    \\
            \hline\hline
            {\bfseries Direct Transfer}      & 71.57               \\
            \hline
            {\bfseries UNIT} \cite{LiuMY01}  & 26.61               \\
            \hline
            {\bfseries DRIT} \cite{Lee01}    & 40.07               \\
            \hline
            {\bfseries CycleGAN} \cite{no_4} & 69.31               \\
            \hline
            {\bfseries VideoGAN}             & { \bfseries 76.69 } \\
            \hline
        \end{tabular}
    \end{center}
\end{table}

The training sets translated by UNIT, DRIT and CycleGAN decrease the segmentation accuracy of polyps, i.e., $-44.96\%$, $-31.50\%$ and $-2.26\%$, respectively, compared to that of direct transfer approach. Since no target domain data are used for training, the content distortion in the translated source domain data deteriorates the accuracy. Oppositely, our VideoGAN can achieve high-quality domain adaptation while maintaining the video contents, which leads to a significant improvement for the Dice coefficient, i.e., $+5.12\%$.


\subsubsection{Data augmentation.} In this experiment, we translate the ETIS-Larib videos to CVC-Clinic domain for data augmentation. ResUnet-50 \cite{He01,Ronneberger01} is adopted as the FCN as well. The FCNs trained with different training sets are evaluated on the additional CVC-Clinic-2018 dataset, consisting of 300 colonoscopic frames\footnote{https://giana.grand-challenge.org/PolypSegmentation/} \cite{VD01}. The polyp segmentation accuracies of FCNs with different domain adaptation methods are listed in Table~\ref{tab2}.

\begin{table}[h]
    \caption{Comparison of Dice coefficients of polyp segmentation yielded by different domain adaptation methods on CVC-Clinic-2018. The best result is in {\bf bold}.}\label{tab2}
    \begin{center}
        \begin{tabular}{l|c}
            \hline
            { }                                         & { \bf Dice (\%)}    \\
            \hline\hline
            {\bfseries No Augmentation}                 & 79.22               \\
            \hline
            {\bfseries Mixing with Original ETIS-Larib} & 79.98               \\
            \hline
            {\bfseries UNIT} \cite{LiuMY01}             & 76.07               \\
            \hline
            {\bfseries DRIT} \cite{Lee01}               & 76.05               \\
            \hline
            {\bfseries CycleGAN} \cite{no_4}            & 81.14               \\
            \hline
            {\bfseries VideoGAN}                        & { \bfseries 83.50 } \\
            \hline
        \end{tabular}
    \end{center}
\end{table}

The FCN trained with the set directly mixing the CVC-Clinic training set and ETIS-Larib videos generates a marginal improvement, i.e., $+0.76\%$, compared to that of baseline, i.e., no augmentation. UNIT and DRIT deteriorate the accuracy due to severe content distortions. Surprisingly, CycleGAN achieves a marginal improvement of $+1.92\%$, probably, because in this data augmentation setting, target domain data outnumbers the translated source domain data and a small amount of distorted source data is ignored by the segmentation network. The proposed VideoGAN yields the highest improvement of Dice coefficient, i.e., $+4.28\%$, compared to the one without augmentations, which remarkably boosts the accuracy of polyp segmentation.


It is worthwhile to mention that the polyp segmentation is a new task announced in GIANA 2018 challenge, which uses the CVC-Clinic-2018 and CVC-Clinic datasets as the training and test set, respectively. As the challenge organizers only announced the team ranking rather than the top segmentation accuracies, we can not directly compare the performance of our approach with state-of-the-art on this dataset.

\subsubsection{Ablation study.} To evaluate the improvement yielded by each component of VideoGAN, an ablation study is conducted on the transfer learning (TF) and data augmentation (DA) tasks. The results of ablation study are presented in Table.~\ref{table1}. The result shows that all of our components can improve the segmentation accuracy -- X-shape generator (X-G) ($+4.99\%$ and $+1.13\%$), color histogram loss ($L_{hist}$) ($+1.56\%$ and $+0.64\%$) and intra-video loss ($L_{i.v.}$) ($+0.83\%$ and $+0.59\%$) -- for the TF and DA tasks, respectively.

\begin{table}[!tb]
    \centering
    \caption{Ablation study of VideoGAN for the transfer learning (TF) and data augmentation (DA) tasks ({\bf Dice (\%)}).}\label{table1}
    \begin{tabular}{c|c|c|c}
        \hline
        {} & Setup                  & {\bf TF} & {\bf DA} \\\hline\hline
        A  & Original CycleGAN      & 69.31    & 81.14    \\\hline
        B  & A + X-G                & 74.30    & 82.27    \\\hline
        C  & B + $L_{hist}$         & 75.86    & 82.91    \\\hline
        D  & C +  $L_{i.v.}$        & 76.69    & 83.50    \\\hline\hline
        E  & A + weight-sharing X-G & 68.24    & 79.20    \\\hline
    \end{tabular}
\end{table}

Furthermore, a variant of X-G, i.e., weight-sharing X-shape generator (X-G), is involved for comparison. The weight-sharing X-G shares weights of the last resiudal block and the first upsampling module -- as same as \cite{LiuMY01} -- for the source and reference frames instead of using dense fusion block. As the weight-sharing X-G easily suffers from the spatial information interference between the paired frames, which may result in the content distortions, degradations of segmentation accuracy ($-1.07\%$ and $-1.94\%$) are respectively observed for the TF and DA tasks, compared to the original CycleGAN.


\subsubsection{Comparison of translated results on CamVid.}
We also evaluate domain adaptation performance of the proposed VideoGAN on the CamVid dataset. The experimental results are presented in { \bf Supplemental Material}.

\section{Conclusion and Future Work}
In this paper, we present a framework, namely VideoGAN, for the domain adaptation of video-based data. To our best knowledge, this is the first work to address the problem of video-to-video domain adaptation. We evaluate the domain adaptation performance of our VideoGAN on the endoscopic and natural videos. The experimental results demonstrate that our VideoGAN can significantly narrow down the gap between different domains.

We plan to improve the performance of our VideoGAN from two aspects. First, the RGB color space is currently used to calculate the color histogram loss. However, we notice that there are some other choices, e.g. CIELab color space, that can provide more accurate color representations compared to RGB color space. Furthermore, the 3D color histogram is also a potential choice for the color histogram loss, replacing the 15-bin histogram. Second, the recurrent neural network (RNN) may be a better solution for the domain adaptation of lengthy videos. However, the whole network may be difficult to train while integrating RNN to GAN, due to its complicated architecture. We will make our best effort to address this problem in the future work.

{\small
\bibliographystyle{aaai}
\bibliography{cr_egbib}
}

\begin{table*}[!htb]
    \centering
    \caption{Semantic segmentation IoU (\%) of the cloudy video from CamVid with VideoGAN.}\label{table_camvid_quantity}
    \small
    \begin{tabular}{l|ccccccccccccc}
        \hline
        {}                & Bicyclist & Building & Car   & Pole  & Fence & Pedestrian & Road  & Sidewalk & Sign  & Sky   & Tree  & mIoU  \\\hline\hline
        \multicolumn{12}{l}{\bf Sunny (validation)}                                                                                      \\\hline
        {PSPNet}          & 78.41     & 88.45    & 83.70 & 22.51 & 65.78 & 53.50      & 92.28 & 82.26    & 29.35 & 91.45 & 81.40 & 68.71 \\\hline\hline
        \multicolumn{12}{l}{\bf Cloudy (test)}                                                                                           \\\hline
        {Direct Transfer} & 4.19      & 35.34    & 35.55 & 8.31  & 2.14  & 22.63      & 54.07 & 45.70    & 9.63  & 5.17  & 37.61 & 23.49 \\\hline
        {VideoGAN}        & 8.98      & 50.44    & 37.24 & 15.27 & 2.84  & 31.62      & 58.33 & 62.63    & 3.45  & 49.68 & 62.77 & 34.18 \\\hline
    \end{tabular}
\end{table*}

\newpage

\section{Appendix}



\subsection{Comparison of translated results on CamVid}
The proposed VideoGAN is also evaluated on the CamVid dataset. The CamVid database contains driving videos under different weathers, e.g., cloudy and sunny. The task translating cloudy videos to the sunny domain is very challenging, as the cloudy videos are often very dark, which loses many detailed information. We translate the cloudy CamVid video to sunny domain with baseline approaches, i.e., UNIT \cite{LiuMY01}, DRIT \cite{Lee01} and CycleGAN \cite{no_4}.

\begin{figure}[!htb]
    \begin{center}
        \includegraphics[width=0.90\linewidth]{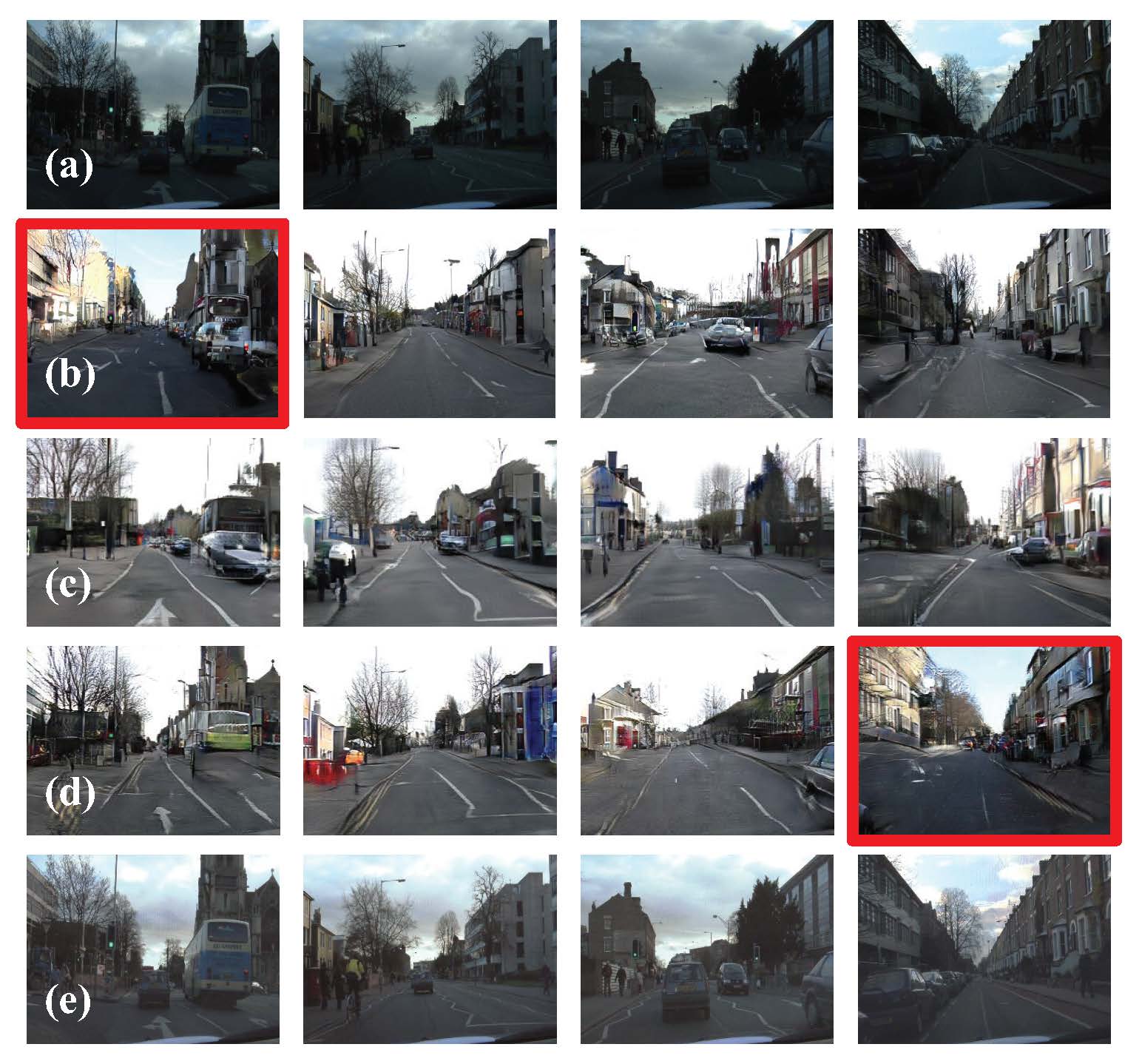}
    \end{center}
    \caption{Domain adaptation results on the CamVid dataset (from cloudy to sunny). Four frames from the begining, middle and ending of the cloudy videos are selected to evaluate the capacity of maintaining intra-video consistency of different translation frameworks. (a) The original cloudy video. (b)-(d) are the translated results produced by UNIT, DRIT and CycleGAN, respectively. (e) The color adaptation result produced by our VideoGAN. {\bf Better viewed on a computer.}} \label{fig6:cross_domains_camvid}
\end{figure}

The translated results are shown in Fig.~\ref{fig6:cross_domains_camvid}. To evaluate the capacity of maintaining the intra-video consistency before and after translation, we extract four frames -- the beginning, middle and ending -- from each of the translated results. The frames marked with red rectangles are found to be mapped to a different mode from that of the other frames. Due to the use of reference frame, acting like an anchor for the whole video, our VideoGAN excellently preserves the intra-video consistency for the translated result. Fig.~\ref{fig6:cross_domains_camvid} also illustrates that, in order to trick on discriminator, the existing image-to-image translation approaches intend to change the image contents. This factor makes the existing unsupervised translation approaches unsuitable for domain adaptation. Our VideoGAN successfully addresses the problem of content distortion, i.e. brightening the cloudy videos while maintaining the contents.

\subsubsection{Semantic segmentation IoU of cloudy videos with VideoGAN.}
The CamVid dataset contains four sunny videos (577 frames in total) and one cloudy video (124 frames). To evaluate the translation performance yielded by our VideoGAN, we train a semantic segmentation network (PSPNet \cite{Zhao_2017_CVPR}) with the sunny videos and test it on the original cloudy video and the translated one. In the experiment, the sunny frames are separated to training (three videos) and validation (one video) sets. The validation (sunny frames) and test (cloudy frames) results are shown in Table~\ref{table_camvid_quantity}. The mean of class-wise intersection over union (mIoU) \cite{Zhao_2017_CVPR} is used for evaluation.

\begin{figure*}[!htb]
    \begin{center}
        \includegraphics[width=0.95\linewidth]{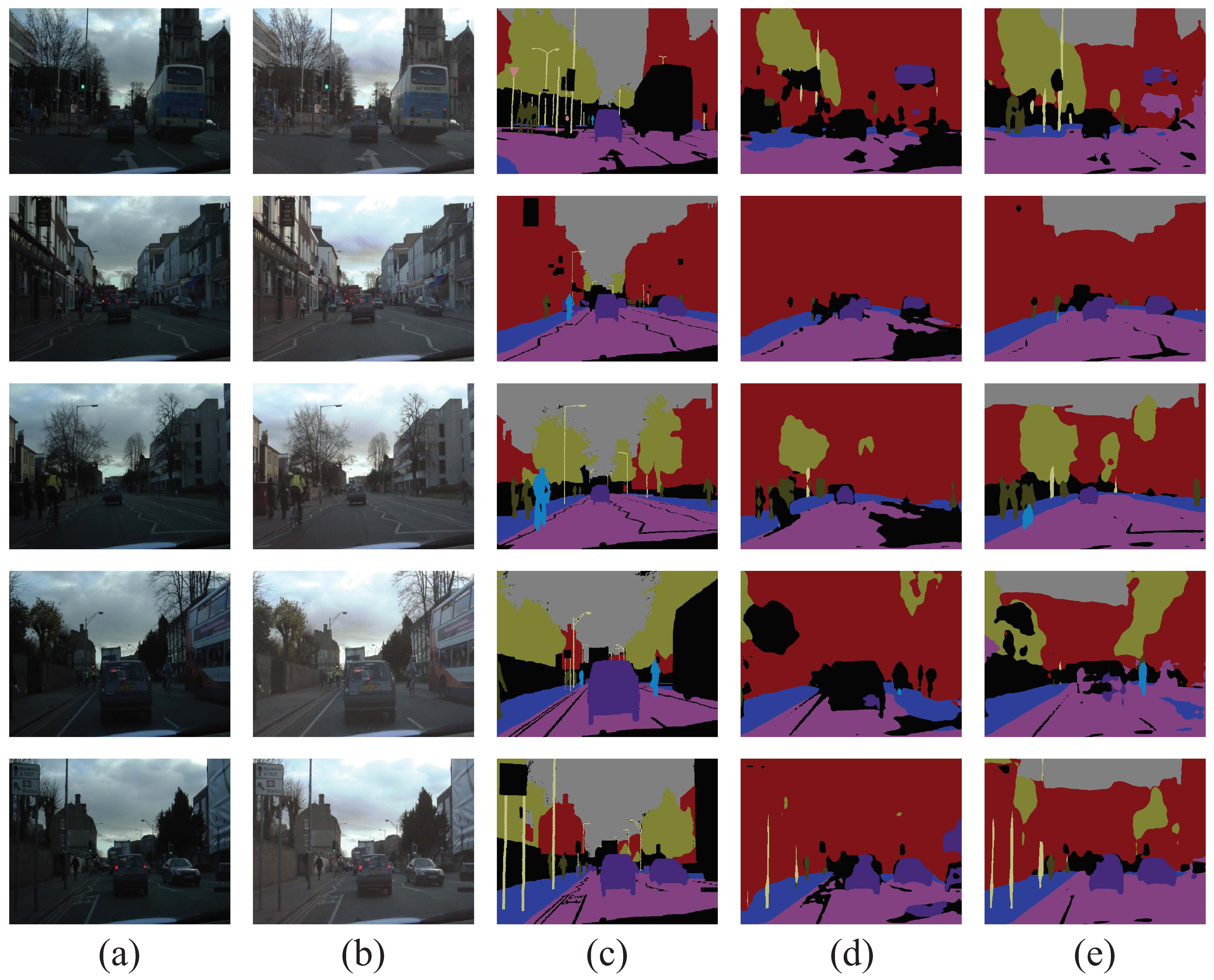}
    \end{center}
    \caption{Segmentation results of cloudy frames with VideoGAN. (a) original cloudy frames; (b) translated results by VideoGAN; (c) ground truth; (d) segmentation results of original cloudy frames; (e) segmentation results of translated frames. {\bf Better viewed on a computer.}}
    \label{fig:seg_res}
\end{figure*}

It can be observed from Table~\ref{table_camvid_quantity} that there is a gap between the validation mIoU (68.71\%) and the test one (23.49\%), which reveals the obvious domain shift between the sunny videos and the cloudy one. Using the proposed VideoGAN, the gap can be narrowed down -- the mIoU of cloudy video is increased to 34.18\% ($+10.69\%$ compared to the direct transfer). For most of the classes, the proposed VideoGAN yields significant improvements such as $+44.51\%$ for Sky, $+25.16\%$ for Tree, $+16.93\%$ for Sidewalk and $+15.1\%$ for Building. The segmentation results with VideoGAN are presented in Fig.~\ref{fig:seg_res}. Fig.~\ref{fig:seg_res} (b) illustrates that our VideoGAN can excellently transfer the cloudy video to sunny while maintains the detailed information such as the plate number ({\bf M734 GPK}) and the pedestrians on the road. {\itshape Please refer to the translated video, which is also submitted as supplemental material, for better view.}

\end{document}